\documentclass[12pt,man]{apa6}
\usepackage{mathptmx}
\usepackage[T1]{fontenc}

\usepackage[style=apa,backend=biber]{biblatex}
\usepackage{threeparttable} 
\usepackage{amssymb,amsmath,amsfonts,amsbsy}
\usepackage{graphicx}
\usepackage{numprint}
\usepackage{tabularx}
\usepackage{setspace}
\usepackage{longtable}
\usepackage{rotating}
\usepackage{censor,caption}
\usepackage{color}
\usepackage{geometry}
\usepackage{listings}
\usepackage{url}
\usepackage{multicol}
\usepackage{tikz}
\usepackage{placeins}
\usepackage{siunitx}
\usepackage{float}
\usetikzlibrary{arrows.meta,positioning,decorations.pathreplacing}
\title{Advantages and limitations in the use of transfer learning for individual treatment effects in causal machine learning}

\author{Şeyda-Betül Aydın \and Holger Brandt}

  \affiliation{Methods Center, University of T\"ubingen}
\shorttitle{}   
\leftheader{}   

\makeatletter
\def\@runninghead{} 

\def\ps@titlepage{%
  \def\@oddhead{\hfill\thepage}%
  \def\@evenhead{\@oddhead}%
  \def\@oddfoot{}%
  \def\@evenfoot{}%
}
\makeatother

\begin{document}

\maketitle

\begin{abstract}
\medskip
Generalizing causal knowledge across diverse environments is challenging, especially when estimates from large-scale datasets must be applied to smaller or systematically different contexts, where external validity is critical. Model-based estimators of individual treatment effects (ITE) from machine learning require large sample sizes, limiting their applicability in domains such as behavioral sciences with smaller datasets. We demonstrate how estimation of ITEs with Treatment Agnostic Representation Networks \parencite[TARNet;][]{shalit2017estimating} can be improved by leveraging knowledge from source datasets and adapting it to new settings via transfer learning \parencite[TL-TARNet;][]{aloui2023transfer}. In simulations that vary source and sample sizes and consider both randomized and non-randomized intervention target settings, the transfer-learning extension TL-TARNet improves upon standard TARNet, reducing ITE error and attenuating bias, when a large unbiased source is available and target samples are small. In an empirical application using the India Human Development Survey (IHDS-II), we estimate the effect of mothers’ firewood collection time on children’s weekly study time; transfer learning pulls the target mean ITEs toward the source ITE estimate, reducing bias in the estimates obtained without transfer. These results suggest that transfer learning for causal models can improve the estimation of ITE in small samples.

\end{abstract}

\newpage


Investigating treatment effects at both the population and individual levels is crucial in many fields, particularly in medical and social sciences. 
The individual treatment effect (ITE) quantifies the difference in an individual’s outcome when receiving a treatment versus not \parencite{jin2023sensitivity}. In contrast to the average treatment effect (ATE), the ITE takes treatment heterogeneity into account and provides person-specific effects given a set of background information (covariates).\footnote{The ITE is often also referred to as Conditional Average Treatment Effect \parencite{shi2024estimating}.}
The ITE is particularly important in personalized medicine or clinical psychology as well as tailored interventions, where the goal is to understand how a treatment will affect each individual rather than just the average effect across a population. 
This information allows for more personalized and effective treatment decisions \parencite{hoogland2021tutorial}.

\textcite{shalit2017estimating} introduced a novel theoretical framework and family of algorithms aimed at estimating the ITE from observational data under the strong ignorability assumption. Their approach, the Treatment-Agnostic Representation Network (TARNet), learns a balanced representation to ensure that treated and control distributions are similar, and they derive a new generalization-error bound that depends on both the standard estimation error and the distance between the induced treated and control distributions.  Using an integral probability metrics (IPM), they establish explicit bounds for distributional distances such as the Wassertein distance and Maximum Mean Discrepancy (MMD). Through experiments on real and simulated datasets, their proposed methods match or outperform state-of-the-art techniques, demonstrating significant potential for causal inference applications in healthcare, economics, and education.

However, a frequent challenge arises from the limited sample size in available datasets, which often stems from the complexity and expense of data collection \parencite{shaikhina2017handling}. This scarcity creates obstacles in accurately identifying causal relationships, especially when using machine learning methods such as TARNet. In general, the strength of machine learning in pattern recognition depends on the amount of data available; smaller datasets consequently reduce both the power and precision of the algorithms \parencite{kokol2022machine}.

The term "small dataset" is context-dependent, varying with the complexity of the task and model. In domains like medicine, even datasets with a few hundred persons may be considered small. \textcite{kokol2022machine} emphasized that small-data situations often arise in areas involving rare diseases or costly experiments, where data collection is inherently constrained. For instance, \textcite{zantvoort2024estimation} observed severe overfitting with fewer than 300 persons in a digital mental health dataset and recommended at least 500–1000 persons for reliable predictions. On the other hand, in large-scale machine learning settings, such as ImageNet with millions of images, even a few thousand examples may be inadequate when training deep neural networks (\cite{rather2024breaking}). In general, a "small dataset" refers to having too few examples to effectively capture the true data distribution, often resulting in poor model generalization.

In social sciences, particularly in clinical psychology, sample sizes tend to be even smaller, thus impacting the reliable use of machine learning methods such as TARNet. 
Meta-analyses on psychological depression treatment and anxiety disorders found an average sample size of $N=106$ or $N=45$, respectively (\cite{Schuster2021,Podina2019}). 
This limits the possibility to estimate ITEs in these fields.


\subsection{Transfer learning}

Causal conclusions rarely remain confined to the dataset in which they were estimated, and extending an effect to a new hospital, region, population, or time period requires addressing external validity: the joint distribution of covariates and outcomes in the target setting often differs from the source \parencite{rojas2018invariant}. Classical fixes such as post-hoc reweighting or stratification help only when all relevant differences are measured and correctly modeled which is an assumption that is hard to guarantee \parencite{huang2024towards}. Simply pooling datasets can also bias estimates if background covariates differ substantially or if treatment effects vary across contexts; when one dataset is much larger, it can dominate estimation and overwhelm signal from the smaller one \parencite{vo2022adaptive,wei2023transfer}. More broadly, an effect that is internally valid in a randomized trial may fail to generalize if the trial’s sample is narrow, while observational data in the target domain may be more representative yet introduce confounding. Ignoring effect modifiers can produce substantial external-validity bias; in some cases, a transferred randomized controlled experiment (RCT) estimate can be more biased for the target population than an adjusted observational estimate \parencite{degtiar2023review}. These pitfalls underscore the risk of naively applying causal conclusions across settings without explicit adjustment.

One promising way to address these challenges posed by small datasets is through transfer learning, which entails applying insights gained in one setting to another (\cite{zhuang2020comprehensive}). The idea is to start with a model trained on a rich dataset which is called as source dataset and adapt it to a small dataset which is called as target dataset, even when the two domains are not identical \parencite{torrey2010transfer}. Transferring causal knowledge is far more complex than standard predictive transfer learning. In causal inference, the goal is to predict how an intervention affects a specific individual’s outcome. 
By searching for representations that capture causal regularities shared across environments, a transfer‑learning approach can absorb stable structures from the source dataset while flexibly adjusting to local idiosyncrasies in the target dataset \parencite{rojas2018invariant}. 
The benefits extend further when multiple datasets record different interventions on the same underlying system.  In such cases, transfer methods can exploit cross‑study invariances, sometimes without needing a complete causal graph, to deliver more reliable predictions in new domains (\cite{magliacane2018domain}).\footnote{\textcite{pearl2022external} formalize conditions under which whole causal Directed Acyclic Graphs (DAGs) can be transferred from one environment to another. This concept of transportability is related to transfer learning but focuses on whole conceptualizations, and not only ITE estimation as we will concentrate on. One of the major differences is that for transfer learning, not the whole causal mechanism needs to be known \parencite{aloui2023transfer}. See more in the discussion section.}

Addressing the open question of how to reuse causal models across studies, \textcite{aloui2023transfer} developed both theory and practice for transfer learning of ITEs using TARNet. They first established a lower bound showing that any transferred model’s performance on a new task is fundamentally limited by the (unobservable) counterfactual error, warning against naïve reuse of source models. Complementary generalization bounds then prove that effective transfer is possible when discrepancies between source and target distributions are sufficiently small. The authors introduce the Causal Inference Task Affinity (CITA) that captures the similarity between source and target dataset. They provide (limited) evidence that the proposed method will improve efficiency in ITE estimation and that source datasets can be judged for its appropriateness to serve for information transfer to a target dataset.
%


\subsection{Scope and Outline}
Despite the widespread application of transfer learning, its potential for improving causal inference remains largely unexplored in social and behavioral sciences. 
This paper examines the conditions under which the transfer learning by \textcite{aloui2023transfer} enhances predictive performance and clarifies causal mechanisms in small-sample settings. We focus on source and target datasets drawn from the same underlying population, with target data that include both randomized and non-randomized interventions, and we explicitly vary the sizes of the source and target samples to see how data availability modulates the benefit of transfer. Across all scenarios we investigated, regardless of whether the target interventions are randomized or biased and across the full range of sample-size configurations, transfer learning reliably conveyed information from source to target and consistently improved model performance.




%

The remainder of the article is structured as follows. We first introduce TARNet and its extension to transfer learning (TL-TARNet), and discuss the underlying assumptions, limitations, and related approaches from machine learning. We then present empirical and simulation results demonstrating the application of the transfer learning framework. Finally, we offer practical guidelines for implementation and outline directions for future research.

\section{Method}

In this section, we describe methods for estimating ITEs and how transfer learning can improve ITE estimation in smaller samples using information from larger ones. We begin by introducing TARNet, then review the approach of \textcite{aloui2023transfer} and related variants.

\subsection{TARNet}

The main quantity in ITE estimation is 
\begin{equation}
\tau(x)=\mathbb E[Y^{1}-Y^{0}\!\mid X=x],
\end{equation}
which represents the difference between a person’s potential outcome under treatment ($Y^{1}$) and their potential outcome under control ($Y^{0}$), given the characteristics of the person that are included in the covariates $x$. Since $\tau(x)$ is not identified without additional assumptions, alternative ways to at least find approximations for this quantity are necessary. Particularly in situations, where randomization to the treatment groups cannot be ensured, additional assumptions are needed.

TARNet and its balanced extension, the counterfactual regression (CFR), estimate ITEs from observational data by learning a representation in which the treated and control groups are distributionally similar. 
Because the same unit can never be observed under both treatment and control, causal inference hinges on making the treated and control groups as comparable as they would be in a randomized trial. 
Simply fitting two separate outcome models, one for treatment group and one for control group, ignores this problem: if certain covariate profiles appear almost exclusively in the treated group, the control model has no information about them and counterfactual predictions become unreliable. 
TARNet mitigates this issue by jointly learning a representation function $\Phi$ and outcome predictions $h$. The representation function $\Phi$ maps the original covariates to a new space in which the distributions of treated and control units are more alike.

To ground their architecture theoretically, \textcite{shalit2017estimating}
first formalize the target quantity which is the expected precision in
estimating heterogeneous effects ($\varepsilon_{\text{PEHE}}$) as squared difference between true ITE and estimated ITE:
\begin{equation}
\varepsilon_{\text{PEHE}}(h,\Phi)
=\int_{\mathcal X}\!\bigl(\hat\tau(x)-\tau(x)\bigr)^{2}\,p_{F}(x)\,dx,
\end{equation}
with 
\begin{equation}
\hat\tau(x)=h(\Phi(x),1)-h(\Phi(x),0).
\end{equation}
%
where \(\Phi\) is the shared representation function and \(h\) refers to the
outcome predictions, which are applied to \(\Phi(x)\) and indexed by
treatment status. 

Since $\varepsilon_{\text{PEHE}}$ involves the counterfactual outcome, it cannot be computed using observed data. 
Instead, two successive bounds on ITE estimation are derived:
\begin{align}
\varepsilon_{\text{PEHE}}(\Phi,h)
&\le
2\bigl(
   \underbrace{\varepsilon_{CF}(h,\Phi)}_{\text{counterfactual loss}}
   + \underbrace{\varepsilon_{F}(h,\Phi)}_{\text{factual loss}}
   - 2\cdot\underbrace{\sigma_Y^{2}}_{\text{variance of } y}
\bigr)\label{eq:pehe1}\\
&\le
2\bigl(
   \underbrace{\varepsilon^{t=0}_{F}(h,\Phi)}_{\text{factual loss } t=0}
   + \underbrace{\varepsilon^{t=1}_{F}(h,\Phi) }_{\text{factual loss } t=1}
   + \underbrace{\operatorname{IPM}_{G}\bigl(
        \{\Phi(x)\}_{t=0},
        \{\Phi(x)\}_{t=1}
     \bigr)}_{\text{distance between groups}}
\bigr).
\label{eq:pehe2}
\end{align}
There are two components in the first bound in Equation~\eqref{eq:pehe1}; 
the first one is the counterfactual loss $\varepsilon_{CF}(h,\Phi)$ that captures the error in predicting outcomes which is unobservable and the factual loss $\varepsilon_{F}(h,\Phi)$ that measures how well the model predicts the outcomes we actually observe in the data. 
The equation implies that $\varepsilon_{\text{PEHE}}$ cannot be small unless both factual and counterfactual losses are small. 

Since we cannot compute the counterfactual loss directly, the second upper bound in Equation~\eqref{eq:pehe2} is derived. This bound includes two major components: First, $\varepsilon_{F}^{t=0}(h,\Phi)$ and $\varepsilon_{F}^{t=1}(h,\Phi)$ are factual losses for the control and treatment groups, respectively. 
Second, an imbalance between treatment groups due to non-randomization is quantified by an Integral Probability Metric (IPM) that measures the distance between the representation spaces \(\{\Phi(x)\}_{t=1}\) (treatment group) and \(\{\Phi(x)\}_{t=0}\) (control group). The subscript \(G\)  indicates a family of functions.  Intuitively, each function \(g\in G\) is like a test that tries to detect a difference between the treated and control representations. The IPM compares all these tests and keeps the largest difference that any of them can find. Therefore, \(\operatorname{IPM}_{G}(p,q)=0\)  only when none of the functions in 
\(G\) can tell the two groups apart that  means the treated and control representations look the same to every function in \(G\).
In practice, the authors choose G to be the set of all 1-Lipschitz functions, and a classical result (Kantorovich–Rubinstein duality) states that when you use this particular \(G\), the IPM becomes exactly the 1-Wasserstein distance, which is a standard measure of how far two distributions are in optimal-transport terms (\cite{villani2009optimal}).

If both factual losses within treatment and control groups (the model predicts the observed outcomes well) and the IPM distance go toward zero, the counterfactual error will be small, too. 
By minimizing the upper bound, the algorithm thus effectively reduces both the observable prediction error and the risk of extrapolation errors due to covariate imbalance.

\subsection{Estimation of upper bound}
The TARNet learns to estimate the potential outcomes by minimizing the following likelihood\footnote{The Derivation of the formula can be found in \textcite{shalit2017estimating}}:
\begin{align}
    \mathcal{L}(\Phi, h) & =\underbrace{\frac{1}{N} \sum_{i=1}^N w_i \cdot \ell_{(\Phi, h)}\left(x_i, t_i, y_i\right)}_{\text{empirical loss}}
    +\alpha \cdot \operatorname{IPM}_G\left(\left\{\Phi\left(x_i\right)\right\}_{i: t_i=0},\left\{\Phi\left(x_i\right)\right\}_{i: t_i=1}\right)
\label{eq:overall_loss}
\end{align}
with
\[
w_i = \frac{t_i}{2\nu} + \frac{1 - t_i}{2(1 - v)}, \quad
v = \frac{1}{N} \sum_{i=1}^N t_i
\]
The first term in Eq.~\eqref{eq:overall_loss} is a sample-weighted empirical loss:  \(\ell_{(\Phi,h)}(x_i, a_i, y_i)\) measures how well the model predicts the
factual outcome \(y_i\), that is, it minimizes the factual losses in Equation~\eqref{eq:pehe2}. The second term minimizes any residual covariate
imbalance after the observations have been mapped through the representation function~\(\Phi\). Intuitively, minimizing this term pushes the two distributions in the representation space closer together. \(v\) indicates how many treated units exist, and \(w_i\)
uses this to weight treated and control examples so that both groups influence the model equally despite imbalance.

The hyperparameter \(\alpha>0\) is a user-chosen balancing weight, which can also be optimized by tuning for the loss against the similarity of the treated and control representations. 
Setting \(\alpha=0\) yields the original TARNet loss, while \(\alpha>0\) results in the CFR (counter-factual regression) objective of \parencite{shalit2017estimating}. 


Taken together, the generalization bound shows that the unobservable error of an ITE estimator is controlled by two measurable training terms: the factual loss and the imbalance term, which measures how similar the treated and control feature distributions become under the learned representation. When the representation makes these distributions nearly indistinguishable, imbalance shrinks; when predictions are accurate, factual loss shrinks. Thus, if both are small, the counterfactual error, and therefore the treatment effect error, is also small. Training TARNet minimizes exactly these observable terms, tightening the bound and reducing the true ITE error $\varepsilon_{\text{PEHE}}(h,\Phi)$.


\subsection{Extension of the TARNet with Transfer Learning (TL-TARNet)}
\textcite{aloui2023transfer} extend the TARNet into the domain of transfer-learning. 
They show that, for a source model applied to a target task, the target $\varepsilon_{\text{PEHE}}(h,\Phi)$ can be upper-bounded by the source factual losses plus three IPM terms that measure mismatch between source and target datasets.
The following equation is defining $\varepsilon_{\text{PEHE}}(h,\Phi)$, which measures how well the model that was trained on the source  ($\hat{f}^S$) predicts individual‑level treatment effects on a new target population:
\begin{equation}
\varepsilon^{T}_{\text{PEHE}}(\hat{f}^S) = \mathbb{E}_{x \sim p^T_F} \left[ \left( \tau^T(x) - \left[ \hat{f}^S(x,1) - \hat{f}^S(x,0) \right] \right)^2 \right]
\label{eq:overall_loss2}
\end{equation}
where the subscripts $T$ and $S$ indicate target and source datasets, respectively. $\tau^T(x) = \mathbb{E}[Y^{T,1} - Y^{T,0} \mid X^T = x] $ is the true ITE in the target task and $\left( \hat{f}^S(x, 1) - \hat{f}^S(x, 0) \right)$  is the model’s ITE prediction obtained by subtracting its two outcome predictions of the source dataset. Because the counterfactual outcome is never observed, the $\varepsilon_{\text{PEHE}}$ cannot be computed directly. However, it can be bounded using quantities that are observable. $\varepsilon_{\text{PEHE}}$  can be bounded in terms of factual and counterfactual loss as follows:  
\begin{equation}
\underbrace{\epsilon_F^T}_{\text{observed error}} + 
u\cdot\underbrace{\epsilon_{CF}^{T, a=0}}_{\text{unobserved error}} \leq 
\underbrace{\epsilon_{\text{PEHE}}^T}_{\text{ITE error}},
\label{eq:overall_loss3}
\end{equation}
where the term $u$ is the fraction of treated individuals in the target task and is used to weight the counterfactual loss.

Earlier transfer‑learning methods for causal inference, such as those of \textcite{vo2022adaptive} and \textcite{bica2022transfer}, addressed only the distributional shift between source and target domains while presuming the same causal mechanism holds in both. In social domains applications, however, this assumption is not always fulfilled; covariate distributions can change, treatment propensities can vary, and the way treatments affect outcomes may differ across contexts. Different from previous studies, \textcite{aloui2023transfer} confront these challenges simultaneously with the following bound for the counterfactual error in Equation~\eqref{eq:overall_loss3}: 
\begin{equation}
\epsilon_{CF}^T(\hat{f}) \leq 
\underbrace{\epsilon_F^S(\hat{f})}_{\text{source fit}} 
+ \underbrace{\text{IPM}_G(p_F^T, p_F^S)}_{\text{distribution difference}} 
+ \underbrace{\text{IPM}_G(p_F^T, p_{CF}^T)}_{\text{treatment-selection bias}} 
+ \underbrace{\mathbb{E}_{(x,a)\sim p_F^S} \left[ \left| f^S(x,a) - f^T(x,a) \right| \right]}_{\text{causal-mechanism change}}
\label{eq:overall_loss4}
\end{equation}
Their upper bound on the target counterfactual loss isolates three distinct contributions -- source‑to‑target distributional shift, treatment‑selection bias within the target dataset, and changes in the causal mechanism between source and target dataset which is making it clear how each source of mismatch inflates the error. 
As a consequence, $\varepsilon_{\text{PEHE}}$ can be written in the following form:
\begin{equation}
\epsilon_{PEHE}^T(\hat{f}) \leq 4\cdot \epsilon_F^S(\hat{f}) + 4\cdot \text{IPM}_G(p_F^T, p_F^S) + 2\cdot \text{IPM}_G(p_F^T, p_{CF}^T)
+ 4 \cdot \mathbb{E}_{(x,a) \sim p_F^S} \left[ \left| f^S(x, a) - f^T(x, a) \right| \right].
\label{eq:overall_loss4}
\end{equation}
The bounding equation of $\varepsilon_{\text{PEHE}^T}$ does not contain the counterfactual loss explicitly; instead, it is bounded by four observable terms. When the source factual loss, the source–target distributional mismatch, the treatment‑selection bias in the target dataset, and the causal‑mechanism discrepancy are all small, the counterfactual loss must also be small, and the overall error remains low.
The following assumptions need to be satisfied to ensure that the causal transfer learning is valid: 
\begin{itemize}
    \item Assumption 1: The representation space $\Phi$ is injective ($\Phi^{-1} \text{ exists on } \text{Im}(\Phi)$) and preserves all causal signal through the encoder, which means that we are preventing any information loss.
    \item Assumption 2: There exists a real function space \( G \) on \( \text{Im}(\Phi) \) such that the function \( r \mapsto \ell_{\Phi,h}^T(\Psi(r), a, y) \in G \). This assumption ensures differences in representation distributions (treatment vs. control, or source vs. target) are a good proxy for differences in the representation‑level loss.
    \item Assumption 3: There exists a function class \( G' \) on \( \mathcal{Y} \) such that \( y \mapsto \ell_{\Phi,h}(x, a, y) \in G' \). It guarantees that differences between source and target outcome distributions translate into proportionally controlled changes in the loss.
\end{itemize}
Supposing that Assumptions 1, 2 and 3 hold, performance of the model which is transferring causal knowledge from the source to the target dataset,  $\varepsilon_{\text{PEHE}}(h,\Phi)$ is upper bounded as follows:\footnote{Derivation of the formula can be found in \textcite{aloui2023transfer}} 
\begin{equation}
\begin{aligned}
\varepsilon_{\text{PEHE}}^{T}(\Phi,h)
\;\le\;
2\Bigl(
      &\;\underbrace{\varepsilon_{F}^{S,t=1}(\Phi,h)
      +   \varepsilon_{F}^{S,t=0}(\Phi,h)}_{\text{factual loss source}} 
      +\,\underbrace{\operatorname{IPM}_G\left(\left\{\Phi\left(x_i^T\right)\right\}_{t=1},\left\{\Phi\left(x^S\right)\right\}_{t=1}\right)}_{\text{distance target-source } t=1}\\
      &+\,\underbrace{\operatorname{IPM}_G\left(\left\{\Phi\left(x^T\right)\right\}_{t=0},\left\{\Phi\left(x^S\right)\right\}_{t=0}\right)}_{\text{distance target-source } t=0} 
      +\,\underbrace{\operatorname{IPM}_G\left(\left\{\Phi\left(x^T\right)\right\}_{t=0},\left\{\Phi\left(x^T\right)\right\}_{t=1}\right)}_{\text{distance between groups in target}}\\
      &+ 2\gamma^{*}
   \Bigr).
\end{aligned}
\label{eq:transfer_bound}
\end{equation}
When the model trained on a source domain is applied to a target domain $\varepsilon_{\text{PEHE}}^{T}(\Phi,h)$ from the target dataset is controlled by five observable quantities. The first two are the factual losses from the source dataset both for the treated and control units ($\varepsilon^{S,t=1}_{F}(\Phi,h), \varepsilon^{S,t=0}_{F}(\Phi,h)$). The next three terms are IPM distances: two compare the treated-treated and control-control distributions across the source and target domains, while the third is identical to the treated-control distance within the target domain in TARNet. 
Finally, the residual $\gamma^{*}$ is defined as the expected IPM between the source and target outcome distributions conditional on~$X$. It captures any irreducible mismatch in causal mechanisms:
\begin{equation}
 \quad
\gamma^* = 
\mathbb{E}_{x \sim p_F^S}
\left[
\operatorname{IPM}_{G'}\left(
P(Y_t^S \mid x),\,
P(Y_t^T \mid x)
\right)
\right]
\end{equation}
Minimizing all all five terms will lead to a successful transfer of the causal mechanisms. The resulting bound naturally reduces to the original TARNet bound when the source and target coincide, but it shows explicitly which divergences must be small for a transfer to succeed.

\subsection{Implementation of $h$ and $\Phi$ in TL-TARNet}

Training begins by optimizing the entire TARNet as one shared encoder $\Phi$ and followed by two outcome heads $h_0, h_1$. A shared multi‑layer encoder first transforms every covariate vector into a latent embedding; on top of that embedding sit two separate heads, one dedicated to predicting the control outcome and the other the treated outcome, so each training example only updates the head that matches its observed treatment arm. During learning the optimizer repeatedly takes mini‑batches, computes the prediction loss on the factual outcomes, re‑weights that loss to correct for any class‑size imbalance, and simultaneously adds a penalty that measures how different the treated and control embeddings are using an IPM such as Wasserstein distance. During each iteration, the factual loss is back‑propagated through both $\Phi$ and $h_0, h_1$ and the process repeats until convergence. In this way $\Phi$ is encouraged to produce a representation where the two treatment groups overlap, while $h_0, h_1$ specialize on outcome prediction. 
\vspace{5mm}

\begin{center}
    \includegraphics[width=\linewidth]{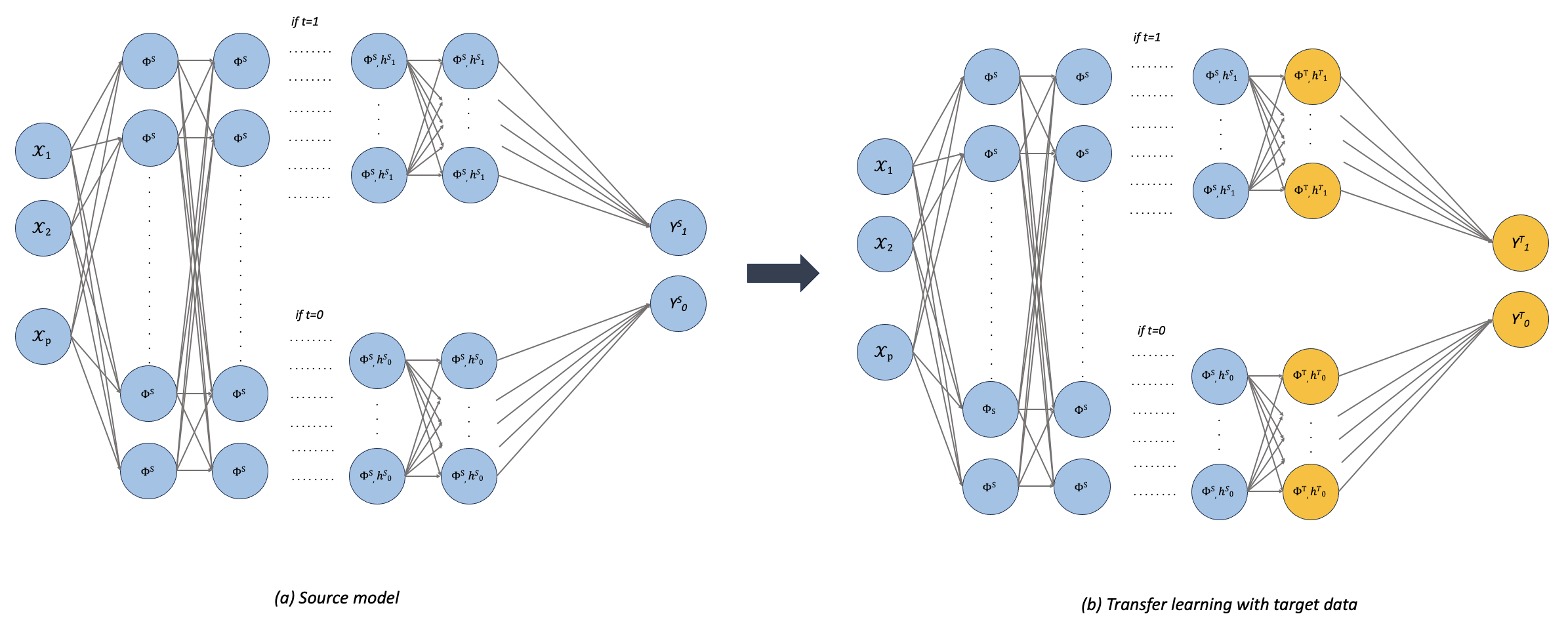}
    \captionof{figure}{Representation of transfer learning.}
    \label{fig:tl1}
\end{center}

Panel (a) of Figure 1 shows the source model once it has been fully trained; a shared encoder $\Phi^S$ (blue circles) transforms the covariates, and two specialized heads $h^S_0$ and $h^S_1$ map that representation to the potential outcomes $Y^{S,0}$ and $Y^{S,1}$.

Panel (b) illustrates the transfer-learning stage. All parameters up to a chosen depth-including the entire encoder and, if desired, part of each head are copied unchanged into the new model and kept fixed (still shown in blue). Only the remaining block (colored in yellow in our example) is re-initialized and left trainable; this lightweight section, denoted $\Phi^T,\,h^T_0,\,h^T_1$, is then fine‑tuned on the smaller target dataset. Freezing the early layers preserves the rich features learned from the source task, while updating the last one or two layers (depending on the target dataset size) lets the model adjust to the specific distribution and label characteristics of the target domain.
\vspace{5mm}

\begin{center}
    \includegraphics[width=\linewidth]{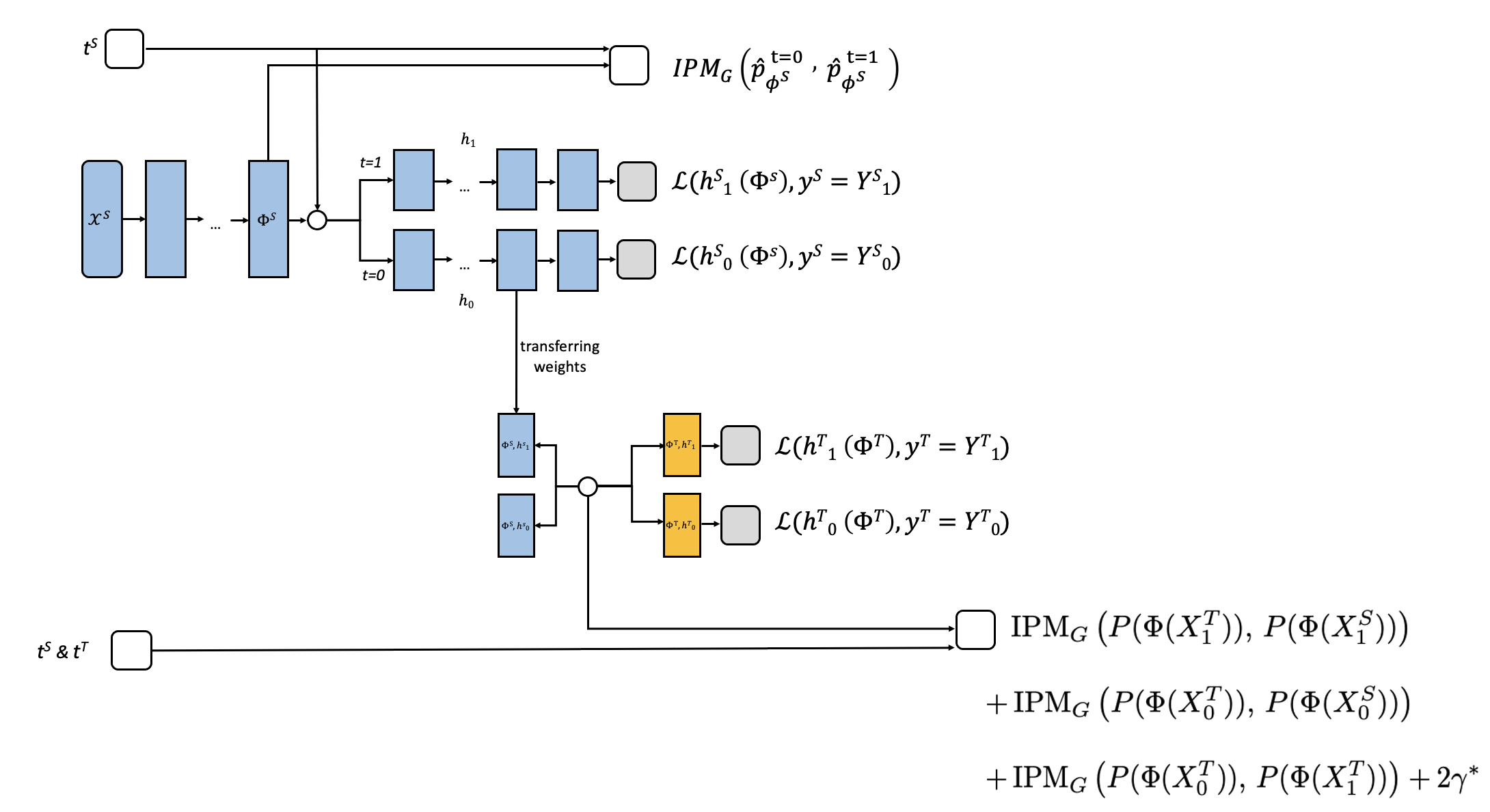}
    \captionof{figure}{Representation of transfer learning with IPMs.}
    \label{fig:tl2}
\end{center}


A detailed version of transfer learning training with TARNet in two stages is illustrated in Figure 2. In the transfer learning stage, the network undergoes fine-tuning using a smaller target dataset. At this part, weights learned from the source dataset are transferred, and training continues exclusively with the smaller target dataset, with an emphasis on reducing discrepancies in representation spaces and minimizing the target dataset factual loss. Specifically, there are three IPM components: (i) aligning treated versus control representations within the target dataset, (ii) aligning treated group representations between source and target datasets, and (iii) aligning control group representations between source and target datasets. Gradients update the encoder $\Phi$, progressively shrinking discrepancies both within and across domains. Simultaneously, outcome heads $h_0, h_1$ adapt to align with target-specific representations. This targeted fine-tuning process typically lasts for only a few epochs and affects selected network layers (usually a few layers), allowing the model to retain generalizable features learned from the source while adapting to the new target task.

\subsection{Causal Inference Task Affinity (CITA)}

When several source datasets are available, the first challenge is to decide which one is worth fine‑tuning for the new task. CITA meets this need by computing affinity between the source and the target datasets before any transfer learning is applied. The CITA score can be used to identify sufficiently similar datasets: sources with a low CITA score are close to the target and empirically yield small transfer error, whereas sources with a high score signal a high risk of negative transfer and should be avoided.  Reliable use of CITA, presumes that the source model has been trained to represent its own task well (\cite{aloui2023transfer}).

After training a TARNet on the source dataset, we keep the learned weights fixed and calculate two Fisher information matrices one using the same source data $F^{S,S}$ and one using the target dataset $F^{S,T}$. For a network $N_{\theta}$ trained on source data $D^S$ and a loss $L(\theta,D)$, the Fisher matrices are then
\begin{align}
F^{S,S}
   &=\,
   \mathbb{E}_{D\sim D^S}
   \bigl[
      \nabla_{\!\theta}L(\theta^S,D)\,
      \nabla_{\!\theta}L(\theta^S,D)^{\!\top}
   \bigr]\\
F^{S,T}
  &=\,
   \mathbb{E}_{D\sim D^T}
   \bigl[
      \nabla_{\!\theta}L(\theta^S,D)\,
      \nabla_{\!\theta}L(\theta^S,D)^{\!\top}
   \bigr].
\end{align}
Each matrix is the expected outer product of the loss gradients with respect to the network’s parameters, providing a weight‑by‑weight measure of sensitivity. 

The task affinity score is then the Frobenius norm of the difference. The Frobenius norm, which is the Euclidean length of a matrix when viewed as a vector, measures the overall magnitude of the discrepancy across all parameters. 
\begin{align}
d^{S,T}=\frac{1}{\sqrt{2}}\bigl\lVert (F^{S,S})^{1/2}-(F^{S,T})^{1/2}\bigr\rVert_{F},
\end{align}
The task affinity is the difference between the principal square‑roots of these matrices and values of $d^{S,T}$ near 0 indicate high similarity between the causal tasks, whereas values near 1 signal marked dissimilarity (additional details on this measure when treatment labels are switched are provided in Appendix~\ref{app:cita}).

\subsection{Advantages and limitations of the approach}
The transfer framework of \textcite{aloui2023transfer} provides a principled way to reuse causal models trained on larger datasets for smaller but similar datasets. It begins with a well-trained network on a source dataset and fine-tunes only on the target sample, which helps control overfitting and variance which are the common pitfalls in small clinical or social science studies. The target error is decomposed into three observable gaps which are covariate shift, treatment-selection bias, and changes in the causal mechanism and the method is designed to shrink all three simultaneously. Prior transfer-learning approaches typically addressed only one of these (distribution shift, treatment imbalance, or mechanism drift) in isolation; \textcite{aloui2023transfer}’s unified treatment is thus especially useful in social sciences and health settings where policy, prescribing, and population mix often change together. Importantly, the upper-bound guarantees rely only on factual data and a frozen source network, so no unobservable counterfactual labels are required which is making the method practical in real-world deployments.
\medskip

CITA complements this framework by offering a fast, pre-transfer measure of task closeness. Its score is label-invariant: by minimizing distance over all permutations of treatment labels, tasks that differ only by a relabeling receive distance 0. Because the score correlates closely with the counterfactual error observed after fine-tuning, practitioners can select the most compatible source dataset in advance and avoid inefficient transfer.

There are also some limitations in the \textcite{aloui2023transfer}'s work. From a practical perspective, estimating diagonal Fisher matrices on both source and target dataset (and on each label permutation) means extra computation on top of transfer learning training and storage of large parameter‑sized tensors, which can be costly in terms of time and computation. 
Secondly; it should be highlighted that the Fisher matrices that drive both the theoretical bounds and CITA must come from source models that are already error accurate on their own tasks. Poorly fitted sources produce meaningless distances and can mislead the selection process. 

\subsection{Extensions and alternative approaches}

During the last about five years, several alternative approaches were developed with a similar aim to transfer information from a source to a target dataset.

\textcite{bica2022transfer} address the challenge of improving ITE estimation in a target domain by leveraging information from a source domain with a different feature space. This heterogeneous transfer learning problem is particularly relevant in fields like healthcare, where treatment effects need to be estimated for new patient populations with different clinical covariates and limited data availability. To tackle this, the authors propose a framework that combines representation learning to align heterogeneous feature spaces with a flexible multi-task architecture. 
They extend standard ITE learners to their transfer learning counterparts and evaluate their approach on a new semi-synthetic benchmark designed for heterogeneous transfer learning. Their results demonstrate significant performance improvements over traditional methods while offering insights into how different transfer strategies affect causal effect estimation across domains. 
Compared to \textcite{aloui2023transfer}, their approach only allows for variations in the covariate distribution, but not for changes in the causal mechanisms.

Similarly, \textcite{aglietti2020multi} construct a joint model that integrates interventional source data with observational target data. They employ the concept of transportability, which lets researchers use observational data from the source population for the target population, even in the absence of randomized trials in the target group. 

Along these lines, \textcite{vo2022adaptive} investigate ways to enhance the estimation of treatment effects within a specific target population. They achieve this by using diverse yet related data sources, while considering the potential differences in the distributions between the source and target populations. Also, they present three levels of knowledge transfer by modeling the outcomes, treatments, and confounders ensuring a consistent transfer. For their approach, they assume that DAGs have to be identical, which is crucial because they focus on observational data only. 

\textcite{wei2023transfer} introduce a procedure based on generalized linear models (GLMs) with penalization for situations with many covariates. They design an approach for scenarios where source and target domains share most of the same covariate space. The proposed method, incorporates regularization with transfer learning for estimating nuisance parameters, such as propensity scores, and integrates these estimates into standard causal effect estimators like inverse probability weighting. 

\textcite{chauhan2024dynamic} propose a HyperITE via soft weight sharing that allows dynamic information exchange between treatment groups. This approach is not a classic transfer learning approach because source and target dataset stem from the same dataset. They use a mirror approach were the experimental group provides the source dataset for the target control group and vice versa. 
Experimental results show that HyperITE reduces ITE estimation error, with its benefits becoming more pronounced as dataset size decreases.
Lastly, it is important to distinguish between transfer learning and transportability, a concept developed in \textcite{pearl2011transportability}.
While both concepts aim at reusing causal knowledge across settings, they differ in in their assumptions. 
Transportability relates to the questions when and how causal conclusions (mechanisms or effects) can be validly moved from one population to another. Here statistical alignment alone is insufficient; external validity is central \parencite{pearl2022external}. Sound transport requires explicit assumptions about which mechanisms are invariant and which change across environments which are often formalized with selection diagrams. Violations of these assumptions can yield biased or invalid conclusions \parencite{bareinboim2012transportability, bareinboim2013general}.

Causal transfer learning, in contrast, seeks to transfer estimated causal effects to a new setting and does not require the entire causal graph to match, as long as the effect of interest remains invariant \parencite{aloui2023transfer}. 
\section{Simulation study} 

In this simulation study, we will provide an evaluation of how well transfer learning of ITEs is applicable to social sciences and typical data characteristics. So far, no such investigation has been conducted. The simulation study will shed light on necessary sample sizes for source and target data as well as investigate how the method can reduce bias when sampling in the target data set was not randomized and violates the ignorability assumption.

\subsection{Data Generation} 
\paragraph{Source Data}

Source data were generated for the following model for the outcome:
\begin{align}
y_i =\alpha + \beta T_i + \boldsymbol\gamma \mathbf{X}_i + \boldsymbol\omega T_i\mathbf{X}_i+\epsilon_i
\end{align}
where $\alpha=0$ was an intercept, $T_i\sim Bern(0.5)$ was a Bernoulli distributed binary treatment variable, and $\mathbf{X}_i\sim MVN(\mathbf{0},\mathbf I)$ included five multivariate normally distributed covariates. $\epsilon_i\sim N(0,\sigma^2)$ was a normally distributed residual. The regression coefficient $\beta=1$ indicated the ATE, $\boldsymbol\gamma$ included the impacts of the covariates on the outcome, and $\boldsymbol\omega$ induced treatment heterogeneity, i.e. the ITE depended on the set of covariates. Note that in the generation of source data, the ignorability assumption ($(Y^1_i,Y^0_i)\perp T_i\mid \mathbf X_i$) is satisfied by construction. 

\paragraph{Target Data}

Based on the source data, we created smaller target data sets by subsampling. 
Under the condition of randomized sampling, we randomly drew $N^T$ persons from the source, thus reflecting the same population.
Under the condition of non-randomized sampling, we actively violated the ignorability assumption:
Persons with a high potential outcome $Y^{0}$ under the control condition were more likely to appear in the treatment arm. Formally, each person received a probability\footnote{\(\operatorname{expit}(z)=1/(1+e^{-z})\) is the logistic link.}  $p_i^T:=P(T_i=1|Y^0_i) = \operatorname{expit}(Y^{0}_i)$.
%
Because $p^T$ increased with $Y^{0}$, the resulting datasets were skewed toward high-$Y^{0}$ treated subjects and low-$Y_{0}$ controls.

\paragraph{Data conditions}

We generated large source data sets $N^S= 1000, 5000, 10000, 20000$ and $30000$ persons.
This range reflected smaller data sets as well as large-scale data sets (as they might be available in open source repositories).

For the target data sets we included sample size of $N^T=50, 100, 250$  and $500$ persons. This included typical sample sizes for smaller experiments, as well as large (quasi)-experimental settings. For each condition $R=100$ data sets were generated.

\subsection{Data analysis}

For the data analysis, we used two approaches: The original TARNet that only used the target data and the extended TL-TARNet that first trained a model on the source data and then analyzed the target data sets with the pre-trained models which were explained above. We employed the same general neural network architecture across all simulations, but we adjusted its complexity depending on the dataset size. Because the number of trainable parameters increases with network depth and width, larger datasets can support more complex architectures, whereas smaller datasets require reduced capacity to avoid over-parameterization.

The TARNet architecture consists of a shared encoder and two separate heads for predicting potential outcomes under control and treatment. The encoder maps the input through three fully connected layers, with rectified linear unit (ReLU) activation functions applied after each layer. This shared representation is then passed to two parallel networks (control and treatment), which model the outcome for the control and treatment groups, respectively. Each head consists of hidden layers followed by a final linear layer that outputs the prediction. We use ReLU in all hidden layers to introduce nonlinearity while keeping gradient-based optimization stable and computationally efficient, which is standard practice for feedforward networks in this setting.

In this design, it is important to calculate the number of trainable parameters. Over-parameterization, where the number of parameters exceeds the number of observations, can lead to overfitting or unstable training. Therefore, we constrained all architectures so that the number of trainable parameters was on the order of one tenth of the dataset size. For example, for a source dataset with $10,000$ observations and 5 input features, we used an encoder with 3 layers of 16 neurons each, and 2 hidden layers with 8 neurons in both the treatment and control heads. In this case, the total number of estimated parameters is  $1,074$, which is suitable for training with $10,000$ observations \parencite[see, e.g.,][]{kavzoglu2003use}. 

We selected the learning rate and number of epochs based on the observed convergence behavior of the training loss. For the source dataset with $10,000$ observations, a learning rate of $0.005$, $700$ epochs, and batch size of 32 resulted in stable convergence in the training. Given the relatively small dataset size, we used a small batch size, which is often preferable for optimization and effective use of limited data.

After training on the source data, transfer learning is performed in two phases. In the first phase, we train the model to minimize the distance between the source and target distributions. In the second phase, we minimize the prediction loss on the target data. In both phases, it is important to choose the learning rates carefully. During the minimization of the distance, we use a small learning rate and a larger number of epochs. For instance, for a non-randomized intervention target dataset with 250 observations and a source dataset with 10,000 observations, a learning rate of \(10^{-5}\) with 1,500 epochs led to convergence in both phases.


The datasets were generated in R, and the transfer learning models were implemented in Python using the Torch library. All data and analysis code are available at the following GitHub repository:
\url{https://github.com/PsychometricsMZ/TL_TARNet}.

\subsection{Simulation Results}
In our study we assessed whether transferring a representation learned on a source task improves ITE estimation on a new target task, and understand how that gain depends on source sample size and
on the presence or absence of selection bias in the target data. We evaluate models along three statistics: the estimated ITE mean,
$\varepsilon_{\text{PEHE}}$ and standard error of the estimated means. We do not report results for the setting in which the source data have 1,000 observations and the target data have 500, because in that case the target data constitute a large proportion of the source dataset. In the results below, the target datasets have an average CITA score of 0.18, whereas the non-randomized (biased) intervention datasets have an average CITA score of 0.24.

%

\paragraph{Average ITE}

\begin{center}
    \includegraphics[width=\linewidth]{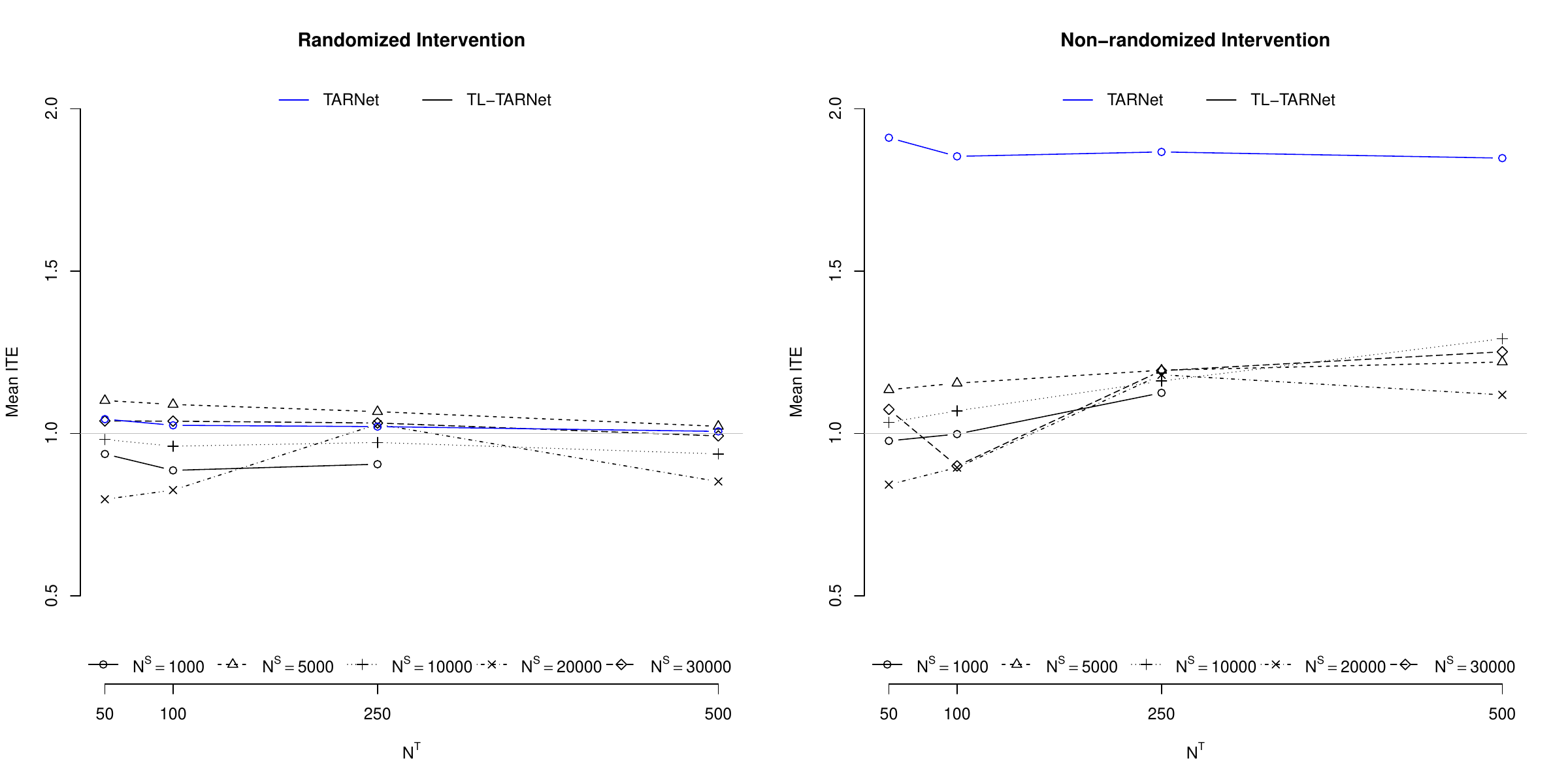}
    \captionof{figure}{Mean ITE for TARNet and TL-TARNet models (i.e. with and without transfer learning) across different sizes of source dataset and target data sets. The population value is indicated with a grey line (${\tau}=1$).}
    \label{fig:sim1}
\end{center}

Figure~\ref{fig:sim1} illustrates the results for the mean ITE estimates. The left panel with randomized intervention shows no bias in the mean ITE, that is also independent of source sample size and target sample size. 

The right panel shows the results for target data sets with violated ignorability assumption. Without transfer learning (TARNet), shows a heavily biased mean ITE close to 2. This bias is not alleviated with increasing target sample size.
For the TL-TARNet, the bias is strongly reduced, indicating that the source dataset (where the ignorability holds) provided an important information that reduced the bias. The bias slightly increases with target sample size (but still substantively below the TARNet). This indicated that increasing target sample sizes overpower the influence of the source model. The bias remains lower if source sample size is higher.


\vspace{5cm}
\paragraph{Precision of ITE estimates}

\begin{center}
    \includegraphics[width=\linewidth]{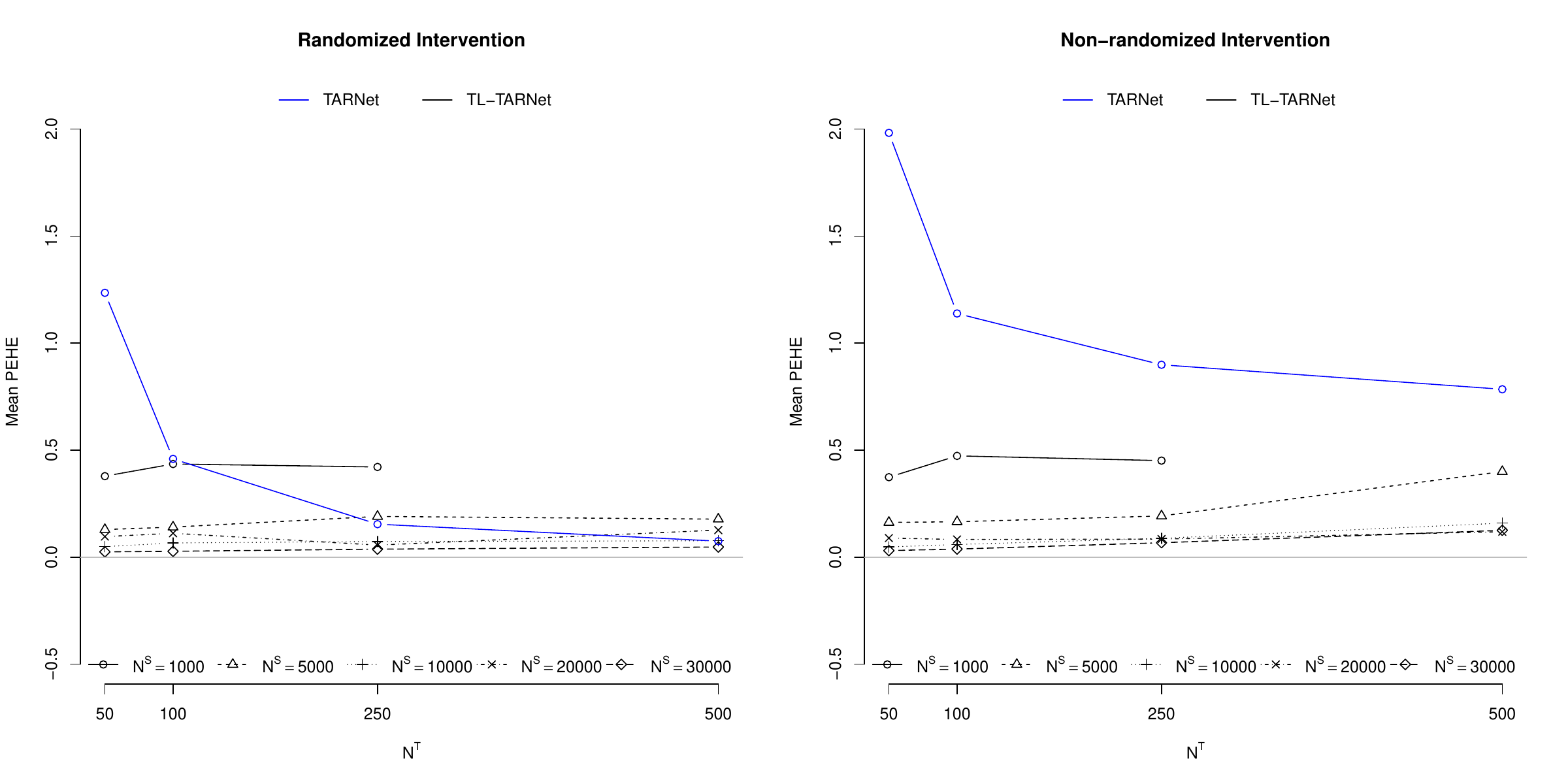}
    \captionof{figure}{Mean $\varepsilon_{\text{PEHE}}$ for TARNet and TL-TARNet models (i.e. with and without transfer learning) across different sizes of source dataset and target data sets. The grey line indicates no error.}
    \label{fig:sim2}
\end{center}

Figure~\ref{fig:sim2} shows the expected precision in estimating ITEs ($\varepsilon_{\text{PEHE}}$). In the left panel with randomized treatment in the target data sets, transfer learning strongly improves this precisions compared to the TARNet, particularly for smaller sample sizes. For target sample sizes of $N\geq 500$, the TARNet provides similar or higher precision than the TL-TARNet. This is due to the higher complexity of the fitted function. For TL-TARNet target sample sizes do not affect the precision. Only the source sample size indicates that $N^S> 5000$ do not substantially improve the precision.

The right panel with the violated ignorability assumptions shows higher levels of imprecision for the TARNet as expected. The TL-TARNet had very similar results for source sample sizes above $N^S=5000$ with slighty higher in the imprecision for larger target sample sizes (in line with the observed bias above).

\paragraph{Summary}

Overall, these results highlight two key insights. First, when the target data are randomized, transfer learning and TARNet perform comparably in terms of bias, though TL-TARNet offers practical advantages in better precision for the ITEs. Second, when the target data are non-randomized, transfer learning substantially improves both bias and estimation precision for ITEs. 

\section{Empirical Study}
To complement the insights gained from the simulation study, we now turn to an empirical application. We use data from the India Human Development Survey (IHDS), Wave II (2011–12), a nationally representative dataset collected by the National Council of Applied Economic Research (NCAER) and the University of Maryland. The survey covers over 42,152 households across 33 states and union territories of India. The IHDS provides detailed information on health, education, jobs, and gender roles to build an overall picture of household welfare across India. 

In this empirical illustration, we build on the framework of \textcite{choudhuri2021lack} and specifically examine their hypothesis, which posits that a greater maternal share of firewood collection time reduces children’s study time. Following the authors, we define the treatment as the mother’s share of total parental time devoted to firewood collection and the outcome as children’s total weekly minutes spent studying (school time plus homework). We control for household income, maternal education, caste–religion group, household electricity access, and distance to school. The sample is restricted to children aged 6–18 who are enrolled in school, in households that cook with firewood and where the mother collects the firewood. Unlike the original paper which additionally includes village development indicators, and state fixed effects, we adopt a more parsimonious specification and exclude these variables.

From the Indian regions in our dataset, we selected Uttar Pradesh as the source dataset because after data cleaning, it had the largest usable sample size ($N^S=1,247$). From our simulation results, this source dataset size deemed feasible. For the target datasets, we considered three scenarios: (i) a random subsample of Uttar Pradesh ($N^T=350$), (ii) a biased subsample of Uttar Pradesh constructed using the same selection function applied in the simulation study ($N^T=350$), and (iii) the Punjab region ($N^T=332$).

While several regions in India could have served as target datasets, we selected Punjab based on both results for the CITA score and training stability. Specifically, Punjab exhibited a lower CITA distance with 0.32 to the Uttar Pradesh source data compared to other regions. For example, we found CITA distances of 0.37 for Jammu \& Kashmir region and 0.33 for Kerala region. Moreover, during model training, the Punjab dataset consistently demonstrated more stable convergence relative to alternative regions. 

For the sampled target datasets (biased and random), we generated 10 candidate data sets for each condition and selected the replication that achieved stable convergence during model training and the lowest CITA score.

We trained the source model on the full Uttar Pradesh data and obtained a mean ITE of $\num{-110}$ minutes per week, with a standard deviation of $7.79$ minutes, implying that increasing mothers’ share of time devoted to firewood collection reduces children’s total weekly study time by nearly two hours on average. 

When transferring to the three target data sets, we expect negative average ITE's consistent with the authors’ findings; specifically, post-transfer estimates should move toward the source mean ($\approx \num{-110}$).

\begin{center}
\captionof{table}{Estimation results of transfer learning using IHDS-II data.}
\label{tab:tl_results}

\small 
\begin{tabular}{lcccccc}
    \hline
    Name & $N^T$  & CITA & Mean with TL & Mean without TL & STD with TL & STD without TL\\
    \hline
    Random-subsample & 350 & 0.09 & -104.31 & -31.86 & 10.48 &  2.03 \\ 
    Biased-subsample & 350 & 0.11 & -94.28  &  25.98 & 16.23 &  4.22\\ 
    Punjab           & 332 & 0.32 & -99.68  &  63.14 & 41.37 &  4.15 \\ 
    \hline\hline
    \multicolumn{7}{@{}l}{\footnotesize
        ($N^T$ = target data sample size; CITA = Causal Inference Task Affinity; 
        TL = transfer learning; STD = standard deviation)}
\end{tabular}
\end{center}

Table \ref{tab:tl_results} shows that Uttar Pradesh's subsamples have a lower CITA than Punjab, which is plausible given they originate from the same population. Given the range of CITA between zero and one, values around 0.10 indicate a high similarity between source and target. 
Without transfer learning, the mean ITEs are positive for the biased subsample and the Punjab region, which was not in line with the hypothesis. The random subsample provides a smaller mean ITE of $\num{-31.86}$ compared to the source data set.
Using the source data and transfer learning, mean ITEs of all targets data sets shift toward the source's ITE with values of $\num{-104.31}$ (random subsample),  $\num{-94.28}$ (biased subsample) and $\num{-99.68}$ (Punjab). These results indicate that ITE estimation was improved for the smaller target data sets by using the source data.

The post-transfer ITE standard deviations are relatively large ($10.48$ for the random subsample, $16.23$ for the biased subsample, $41.37$ for Punjab). Overall the results indicate that transfer learning successfully aligns target estimates with the source in this empirical setting.

\begin{center}
    \includegraphics[width=\linewidth]{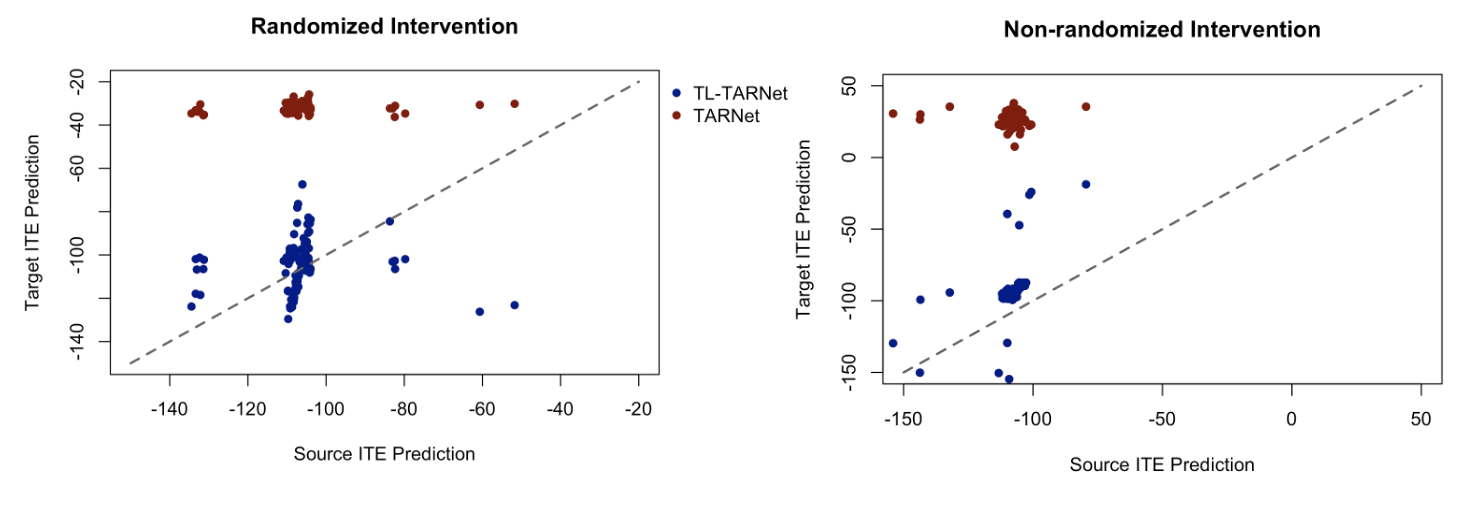}
    \captionof{figure}{Source versus target ITE predictions for randomized (left) and non-randomized (right) intervention settings, with and without transfer learning.}
    \label{fig:emp1}
\end{center}

Figure~\ref{fig:emp1} compares individual-level ITE predictions from the source model (horizontal axis) with the corresponding predictions from the target datasets (vertical axis), under both randomized (left panel) and non-randomized (right panel) intervention settings. Each point represents one individual, with blue points showing predictions obtained using transfer learning and orange points showing predictions from the TARNet model.
The TARNet model produces ITE estimates with very low variance. This is visible from the fact that nearly all orange points lie in a tight horizontal band, regardless of their source ITE values. Moreover, the orange points often lie far from the source predictions shown on the x-axis, revealing a substantial disagreement between the TARNet model on the target data vs. the source data.
When transfer learning is applied, the target predictions exhibit much greater variability across individuals, which appears in the figure as a wider vertical spread of blue points. This increase in variance reflects the recovery of heterogeneous treatment effects that the small target samples alone cannot identify. The dashed diagonal line denotes perfect agreement between source and target ITE predictions, and the systematic deviations from this line, particularly in the non-randomized panels, illustrate how transfer learning corrects the biased target ITE estimates.

\section{Discussion}
In this article, we explored how a transfer learning framework based on the TARNet architecture, following the approach proposed by \textcite{aloui2023transfer}, can be applied to social and behavioral sciences. Our aim was to leverage information from a large source dataset to improve ITE estimation in smaller target datasets. We provided a detailed conceptual and theoretical justification for this framework, emphasizing how transfer learning can mitigate issues related to limited sample sizes and potential selection bias as they are prone in social and behavioral sciences. The effectiveness of the approach was evaluated both through simulation experiments and an empirical data application. Results consistently demonstrated that the transfer learning model was able to improve performance when compared to models trained solely on the small target dataset. These findings underscore the practical value of transfer learning in fields such as health and social sciences, where researchers often rely on relatively small or heterogeneous datasets. By drawing upon a larger, related dataset, it becomes possible to stabilize model estimation and reduce bias, ultimately enabling more reliable inference and prediction even in data-constrained environments.

\subsection{Practical considerations for the application of transfer learning}
Successful transfer learning depends critically on the source model learning a stable, informative representation. Achieving such a learning requires careful monitoring of the optimization process. One major impact of the representation of the source data set is the target-dataset selection. The CITA measure is evaluated based on the trained source model; accordingly, the suitability of a target dataset is assessed in light of this representation. If this initial training was suboptimal, the distance will not represent a meaningful tool that can be used to improve estimation in the target data set.
\medskip

In standard deep learning, a single loss (e.g., MSE) is minimized. With TARNet under transfer learning, optimization becomes multi-objective: the training loss combines prediction error with a distributional distance between treatment and control (or source and target) representations. These terms often differ in scale and evolve at different rates during training. If the hyperparameter $\alpha$ that balances loss and distributional similarities is too small, distribution matching is neglected; if it is too large, the model underfits outcomes to satisfy the distance term. Even with a reasonable $\alpha$, using one learning rate to reduce both components can be difficult because their gradients may conflict in direction or magnitude. Consequently, convergence depends critically on the choice of $\alpha$ and the learning rate. In practice, training may begin with a small learning rate to stabilize optimization (however with higher computational cost), with schedules adjusted after inspecting the loss values. In case of using small learning rates, it is important to increase epoch size substantially. Alternatively, a stepwise procedure can be used: first learn a latent space that aligns the treatment and control groups, and then fit the outcome model by minimizing a predictive loss \parencite[cf.][]{du2021adversarial}.
\medskip

Another consideration during training is the choice of batch size. For small datasets, smaller batches often yield more stable generalization, whereas very large batches can degrade learning dynamics. In TARNet, where the objective typically includes a distributional distance between treatment and control representations, this term is well defined only when both groups are present within the same mini-batch. With group imbalance and very small batches, some mini-batches may contain observations from a single group, leading to noisy distance estimates and instability. Furthermore, before training, the dataset is partitioned into training and test sets, reducing the number of observations available for model fitting. Mini-batch training further lowers the effective sample size per update (and per group in TARNet). Under these conditions, highly overparameterized TARNet architectures such as with many layers or large hidden widths are prone to overfitting and unstable optimization. A more compact architecture 
is therefore generally more appropriate in small-sample settings for both source and target datasets. It is important to note that, researchers can decide how many layers are needed to be frozen and trainable for the target data set. While we used one unfrozen layers in our simulation and empirical application, it is possible to leave more than one layer to be trainable. Increasing the numbers of unfrozen layers naturally will require larger sample sizes in the target data set.

\subsection{Future directions}
Future research should further investigate how to evaluate the similarity between source and target data sets when covariates reflect the same underlying construct but are measured in different ways. An open challenge is how to relate variables that are conceptually similar but not directly comparable. For example, in psychology, different instruments can be used to assess the same construct (such as two different depression scales), yet their scores are not interchangeable. Analogous situations arise in many applied settings, including the present context, where covariates may capture similar dimensions of behavior, environment, or status, but differ in how they are defined or recorded. 

Another direction for future work concerns the merging of multiple data sets to create a big source dataset, particularly when some subgroups or conditions are absent or underrepresented in certain data sources. Addressing these issues is crucial for improving the robustness and generalizability of transfer learning approaches in heterogeneous data environments.

\section{Acknowledgement}

This research was funded by the German Research Foundation (DFG) under grant BR 5175/2-1. The authors also acknowledge support from the state of Baden-Württemberg through bwHPC and from the DFG through grant INST 35/1597-1 FUGG.

\printbibliography
\newpage
\appendix
\section{Additional information on the CITA measures}
\label{app:cita}
Causal inference tasks possess an inherent symmetry: if we swap the treated and control labels, the two outcome-prediction subproblems remain identical except for the sign of the treatment effect. An affinity measure should therefore assign a distance of zero to a task and its label-flipped counterpart. Following \textcite{le2021task}, we compute the one-sided distance $d[s,t]$ (source Fisher curvature on source versus target dataset) and then repeat this calculation after every permutation of the treatment labels in the target set. We denote by $S_{M+1}$ the set of all permutations of the treatment labels $\{0,\ldots,M\}$. For a source-trained network $N_{\theta_a}$, let $F_{a,a}$ be its Fisher information matrix on the source data $D_a$, and let $F_{a,\sigma(t)}$ denote the Fisher matrix of the same network evaluated on the target dataset $D_t$ after the labels have been permuted according to $\sigma \in S_{M+1}$.
For each permutation define

\[
d_{\sigma}
\;=\;
\frac{1}{\sqrt{2}}\,
\bigl\lVert
F_{a,a}^{1/2}-F_{a,\sigma(t)}^{1/2}
\bigr\rVert_{F},
\]
where $F^{1/2}$ is the principal matrix square‑root and $\lVert\cdot\rVert_{F}$ the Frobenius norm.  
The CITA is defined as the minimum of these distances, making the score strictly label‑invariant while requiring no retraining: the two heads of the frozen network are simply read in the opposite order when labels are swapped. CITA distance between the source task $T_{s}$ and the target task $T_{t}$ is
\[
d_{\text{sym}}[s,t]
\;=\;
\min_{\sigma\in S_{M+1}} d_{\sigma}.
\]

\textcite{aloui2023transfer} show that this symmetrized score dominates the one‑sided alternative; CITA drops to zero when tasks differ only by a relabeling and preserves the relative ordering of source tasks under any label flip, whereas the non-symmetrized distance does not. Consequently, CITA provides a more reliable gauge of true task similarity and a safer basis for selecting sources to transfer from.

\end{document}